\documentclass[conference]{IEEEtran}
\IEEEoverridecommandlockouts
\usepackage{cite}
\usepackage{amsmath,amssymb,amsfonts}
\usepackage{algorithmic}
\usepackage{graphicx}
\usepackage{subcaption}
\usepackage{wrapfig}
\usepackage[utf8]{inputenc}
\usepackage[export]{adjustbox}
\usepackage{textcomp}
\usepackage{xcolor}
\def\BibTeX{{\rm B\kern-.05em{\sc i\kern-.025em b}\kern-.08emT\kern-.1667em\lower.7ex\hbox{E}\kern-.125emX}}



\makeatletter
\newcommand*\titleheader[1]{\gdef\@titleheader{#1}}
\AtBeginDocument{%
  \let\st@red@title\@title%
  \def\@title{%
    \bgroup\normalfont\large\raggedright\@titleheader\par\egroup
    \vskip1.5em\st@red@title}
}
\makeatother

\title{A Real-Time Wrong-Way Vehicle Detection Based on YOLO and Centroid Tracking

\thanks{This is funded by ICT incubator project in Chittagong University of Engineering and Technology under the ICT Division of Ministry of Information and Communication Technology, Bangladesh.}
}

\titleheader{2020 IEEE Region 10 Symposium (TENSYMP), 5-7 June 2020, Dhaka, Bangladesh}

\begin{document}

\author{\IEEEauthorblockN{Zillur Rahman\textsuperscript{1}, Amit Mazumder Ami, Muhammad Ahsan Ullah}
\IEEEauthorblockA{\textit{Department of Electrical and Electronic Engineering} \\
\textit{Chittagong University of Engineering and Technology}\\
Chittagong-4349, Bangladesh \\
\{zillur991@gmail.com\textsuperscript{1}, u1502024@student.cuet.ac.bd, ahsan@cuet.ac.bd\}}}

\IEEEoverridecommandlockouts
\IEEEpubid{\makebox[\columnwidth]{978-1-7281-7366-5/20/\$31.00~\copyright2020 IEEE \hfill}
\hspace{\columnsep}\makebox[\columnwidth]{ }}

\maketitle
\IEEEpubidadjcol

\begin{abstract}
Wrong-way driving is one of the main causes of road accidents and traffic jam all over the world. By detecting wrong-way vehicles, the number of accidents can be minimized and traffic jam can be reduced. With the increasing popularity of real-time traffic management systems and due to the availability of cheaper cameras, the surveillance video has become a big source of data. In this paper, we propose an automatic wrong-way vehicle detection system from on-road surveillance camera footage. Our system works in three stages: the detection of vehicles from the video frame by using the You Only Look Once (YOLO) algorithm, track each vehicle in a specified region of interest using centroid tracking algorithm and detect the wrong-way driving vehicles. YOLO is very accurate in object detection and the centroid tracking algorithm can track any moving object efficiently. Experiment with some traffic videos shows that our proposed system can detect and identify any wrong-way vehicle in different light and weather conditions. The system is very simple and easy to implement.
\end{abstract}


\section{Introduction}
\begin{figure}[htbp]
\centerline{\includegraphics[width=8cm,height=5cm]{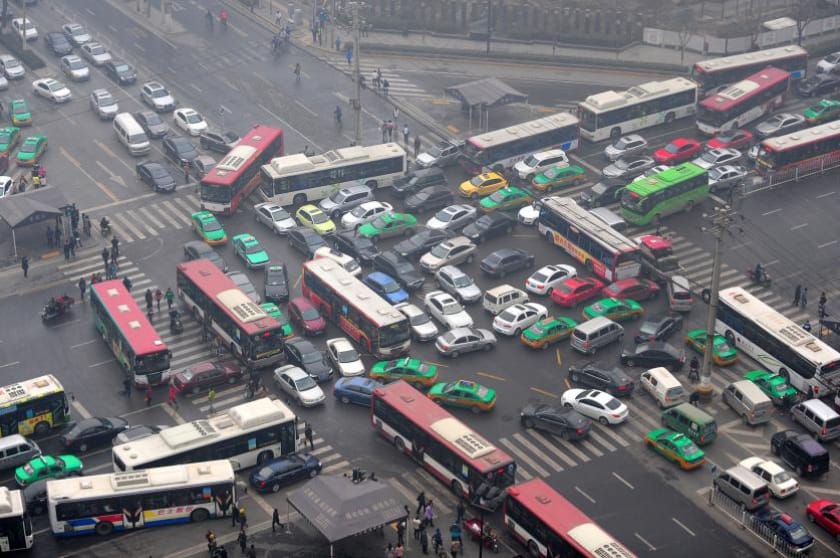}}
\caption{Heavy traffic jam due to wrong-way driving\cite{b4}}
\label{fig01}
\end{figure}

Road accident is a very common issue in a densely populated country like Bangladesh. In 2019, there were 4702 road accidents resulting in 5227 deaths including many children and students\cite{b1}. The capacity of the roadway is not sufficient for the growing number of vehicles and thus imbalance is created. The drivers do not follow traffic rules and take advantage of driving in the wrong-side in times of red traffic signals. It increases traffic on one side and hampers traffic flow greatly \cite{b2}. It also increases the possibility of head-on collision several times. About 355 people die every year due to the crashes in wrong-way driving in the United States \cite{b3}. Fig. \ref{fig01} shows the traffic jam caused by wrong-way driving. This is a common scenario in most of the cities of Bangladesh. So, it is essential to stop drivers from driving on the wrong side. To ensure it, those who don’t follow traffic rules need to find out and strict law should be applied.

Computer vision is used in Intelligent Transportation Systems (ITS) since the beginning of this technology. It is very useful in the traffic surveillance system. The video footage from the on-road cameras can be analyzed to find out the cause of road accidents, traffic jam and also traffic rules violations, counting vehicle and speed. In the past times, it was very difficult to use this technology due to the expensive computation and unavailability of data. But, the recent improvement of different machine learning and deep learning algorithm and availability of powerful Graphics Processing Unit (GPU) and the cheaper camera has made the whole system more efficient. 

The system we present here has three stages. In the first stage, every vehicle in the video frame is detected using the YOLO object detection algorithm and a bounding box is generated for each detected vehicle. Then, the bounding boxes are fed to the centroid based moving object tracking algorithm. The algorithm tracks each vehicle independently in a specified region of interest (ROI). Finally, the direction of the vehicle is determined by calculating its centroid’s height in each frame and detect whether it moves in the wrong direction or not. If the vehicle is on the wrong side, then the system will capture an image of the vehicle.

In the next section, the previous works in this field are described. In section III, the details of the YOLO algorithm are described. In section IV, the details of our proposed system are described. The results are presented in section V and section VI describes the conclusion.

\section{RELATED WORKS}
Computer vision technology is being used in many intelligent traffic monitoring systems. So far, many systems have been developed for wrong-way vehicle detection. The existing methods can be classified into sensor-based and motion pattern-based. Sensor-based methods \cite{b5} use a magnetic sensor to detect the direction of vehicles. The earth's magnetic field gets disturbed in the presence of a vehicle and the sensor gives the signal according to the variation of the magnetic field. From the signal, the direction of the vehicle can be detected. But in this system \cite{b5}, the magnetic field can be varied by other reasons too. In that case, it won’t give a satisfactory result. The motion pattern-based method uses optical flow measurement. The method proposed in \cite{b6} uses optical flow calculation to detect the direction and compare it with the modeled lane direction. But this method faces a problem due to occlusion. Improved optical flow estimation \cite{b7} was used and the authors used background subtraction and Lucas-Kanade method to detect the wrong-way vehicles. But this system is not going to provide a satisfactory result in different light conditions because the background subtraction method is highly affected by the shadow of the vehicle itself.

\section{YOLO}
There are a lot of methods available to detect the vehicles with bounding boxes. Among those methods, for real-time applications, Mask-RCNN \cite{b8} and YOLO\cite{b9} are most popular. The Mask-RCNN method creates bounding boxes using a process called selective search. But for real-time application, YOLO provides more speed than Mask-RCNN and it is the state of the art. There are several versions of YOLO. In this paper, we used the most recent YOLOv3 \cite{b10}. The main thing about YOLO is that it looks at the image once and derives each object with class probability and bounding box and that is why it is very fast comparing with other object detection algorithm. First, it divides the images into the SxS grid cell. Each cell can have B bounding boxes \cite{b9}. If there are C number of classes then, the total bounding box will be SxSxB and each box will have (5+C) attributes. Four attributes for the box coordinate and one is for confidence score that this box contains one of the C classes. There is a threshold confidence score. For our system, we used 0.5 as the threshold. So, it will only show the box with a confidence score of more than 0.5 with the name of the class. All of these operations are completed in a single pass through the network. There is a problem when multiple detections of the same object happen. To overcome this, a process called non-max suppression is used. In this process, first, a box with the highest confidence score is selected. Then, the boxes which have overlap more than a threshold value with the previously selected box will be removed. This process ensures only one box per object.

\begin{figure}[t]
\centerline{\includegraphics[width=8cm,height=5cm]{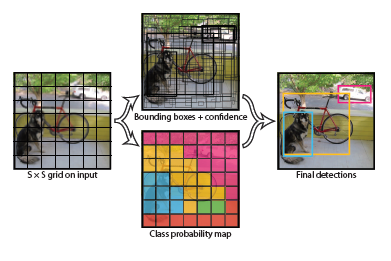}}
\caption{YOLO divides the image into the SxS grid and predicts B bounding box for each cell \cite{b9}}
\label{fig02}
\end{figure}
In YOLOv3, the authors present a new deeper convolutional neural network (CNN) to extract the features. It has 53 convolutional layers. Batch norm and Leaky ReLU activation function were applied to each convolutional layer. The features of the input image are extracted by these convolutional layers. No softmax layer was used because it was not necessary. For the class prediction, binary cross-entropy was used during the training. 

YOLOv3 is trained on the COCO dataset which has 80 different classes. For our system, we only used different classes of vehicles including motorbike, bus, truck, and cars. The previous versions of YOLO did not give good results for small objects but YOLOv3 shows very good results in that case.

\section{PROPOSED METHOD}
This section describes the details of the whole proposed system. The details of each stage will be described consecutively in the following subsections. 
\subsection{Virtual Region of Interest}
\begin{figure}[h]
\centerline{\includegraphics[width=8cm,height=6cm]{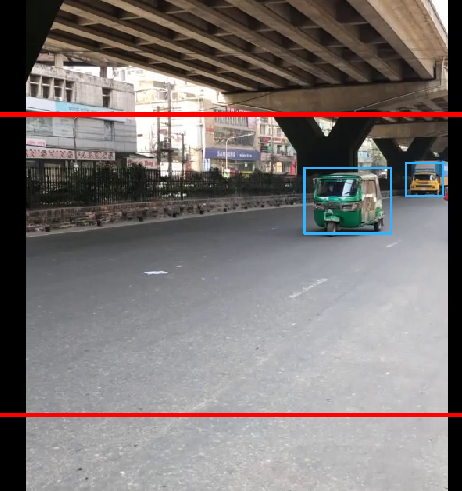}}
\caption{The region of interest for vehicle tracking }
\label{fig03}
\end{figure}

We want to track the vehicles in a certain region of the video frame. That is why we selected a rectangular area in the frame which is shown in Fig. \ref{fig03} in red color. When the vehicle is in the region, it will be tracked. When the vehicle is out of this region, it will be removed from the tracked vehicle list. As this region depends on the camera view, it must be done manually. The region should have a reasonable physical distance or there will be a problem to track the large vehicles.

\subsection{Vehicle Tracking}
\begin{figure}[b]
    \begin{subfigure}{0.25\textwidth}
        \includegraphics[width=4.25cm, height=5cm]{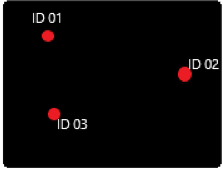} 
        \caption{}
        \label{fig04a}
    \end{subfigure}
    \begin{subfigure}{0.15\textwidth}
        \includegraphics[width=4.25cm, height=5cm]{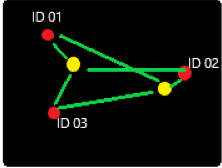}
        \caption{}
        \label{fig04b}
\end{subfigure}
\caption{(a) The first frame (b) second frame}
\label{fig04}
\end{figure}

\begin{figure}[h]
    \begin{subfigure}{0.25\textwidth}
        \includegraphics[width=4.25cm, height=5cm]{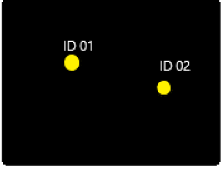} 
        \caption{}
        \label{fig05a}
    \end{subfigure}
    \begin{subfigure}{0.15\textwidth}
        \includegraphics[width=4.25cm, height=5cm]{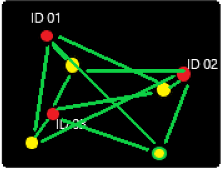}
        \caption{}
        \label{fig05b}
\end{subfigure}
\caption{(a) updated frame (b) when a new frame has more object}
\label{fig05}
\end{figure}

To track each vehicle, we use the centroid tracking algorithm. This algorithm takes the bounding box as the input. So first, the bounding boxes are generated using YOLO. Then, those boxes are fed to the centroid tracker. When the center of each vehicle that means the center of the corresponding bounding box enters the region of interest, it is given a unique identification number which is shown in Fig. \ref{fig04a}. In the next frame, the center of all the objects move in another place or maybe not have any movement which is shown in Fig. \ref{fig04b}. The centroid tracking algorithm is based on an assumption which is that each object will move very little in between the subsequent frame. So, if we can relate any new centroid which has the minimum distance with an old centroid, we can say that this object is previously identified and the new centroid of that object will be updated. This is shown in Fig. \ref{fig05a}. To do this, all possible Euclidean distance between each pair of the new centroids (yellow color) and the old centroids (red color) are computed. The Euclidean distance d between two pixels $(x_i, y_i) and (x_j, y_j)$ is as follow:
\begin{equation}
d=\sqrt{(x_i-x_j)^2 + (y_i-y_j)^2}\label{eq}
\end{equation}

 If the current frame has fewer objects than the previous frame, which is shown in Fig. \ref{fig04b}, that means one or more objects are disappeared from the region of interest. As their centroids are not in the region of interest, the ID of those vehicles will be removed from the tracked vehicle list.

But, when the number of objects in the current frame becomes more than the previous frame which is shown in Fig. \ref{fig05b} then there must be a new object. The algorithm will be updated and old objects will be assigned new centroids. The remaining objects will be identified as new objects and given a new identity number. 

In this way, our proposed system tracks each vehicle independently which is shown in Fig. 06. The figure shows that a CNG vehicle is in the region of interest. The algorithm successfully tracked that vehicle and gave an identity number. On the right side of the image, it is seen that a pedestrian is detected by the YOLO but is not tracked. Because, we feed only the vehicle classes to the centroid tracker and remove others.

\begin{figure}[h]
\centerline{\includegraphics[width=8cm,height=6cm]{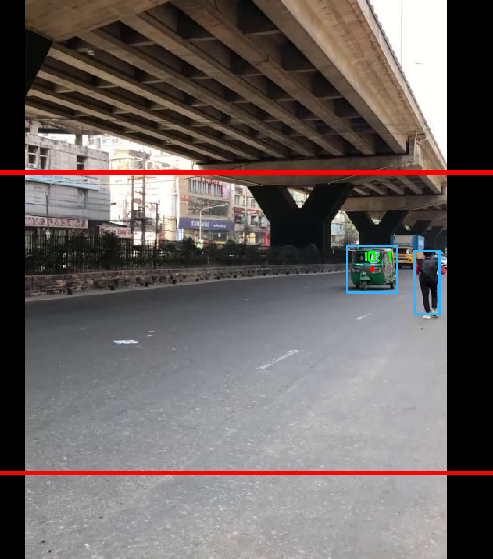}}
\caption{A CNG vehicle is tracked }
\label{fig06}
\end{figure}

\subsection{Wrong-way vehicle detection}
The final step of our proposed system is to detect the wrong-way vehicle. Our system already can track each vehicle that is in our specified region of interest. The centroid of every tracked vehicle has a height from the top of the frame. When the vehicle is registered first and given an identity number, the height of the centroid $H_1$ is computed and stored in a file corresponding to the identity number.

\begin{figure}[htbp]
\centerline{\includegraphics[width=8cm,height=10cm]{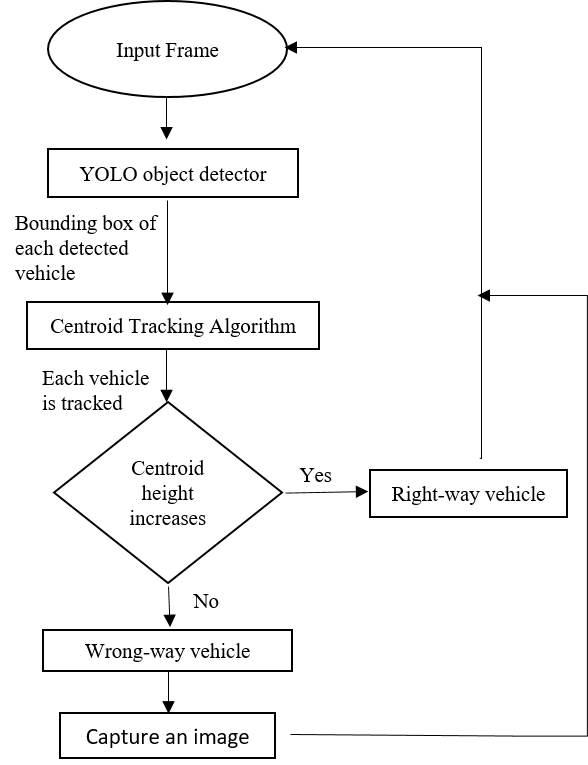}}
\caption{The flowchart of the whole system }
\label{fig07}
\end{figure}

In the next frame, the height of the centroid $H_2$ is computed and stored in another file along with its identity number. This $H_2$ will be updated in each consecutive frame. If the vehicles move, the $H_1$ and $H_2$ of a vehicle will not be equal. By comparing these two heights, our system will predict the direction of the vehicle. In our system, we defined that if the vehicle moves away from the camera, it will be detected as a wrong-way vehicle. So, if $H_1 < H_2$ then, the vehicle is coming towards the camera and is in the right way. If otherwise, our system will detect it as a wrong-way vehicle. The opposite can also be defined just by changing the condition. After the detection of such a vehicle, an image of the frame will be captured automatically for further inspection. The flowchart of the whole system is shown in Fig. \ref{fig07}.

\section*{RESULTS}
\begin{figure}[t]
\centerline{\includegraphics[width=8cm,height=7cm]{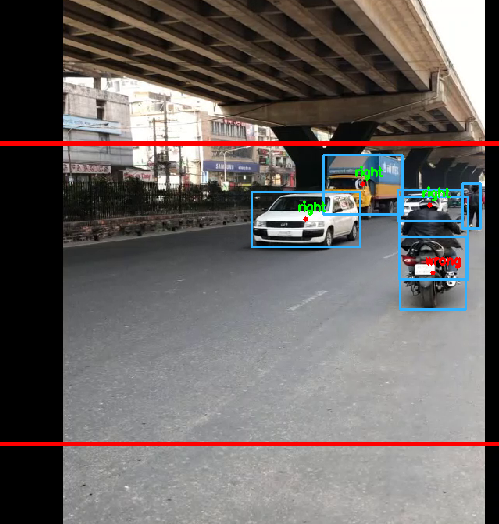}}
\caption{A motorbike is driving on the wrong way}
\label{fig08}
\end{figure}

To check our system, we captured three videos from Chittagong city of Bangladesh each including only one wrong-way vehicle. The videos were captured from roadside and resolution was 1280 x 720 pixels. In Fig. \ref{fig08}, it is seen that our system successfully detected a wrong-way vehicle. Each wrong-way vehicle from three videos is successfully detected. So, the accuracy of the system is almost 100\%. As the other related systems \cite{b5} \cite{b6} \cite{b7} are not based on deep learning algorithm, it is not possible to examine them with our used dataset and experimental setup and also the authors of those work did not mention the accuracy of their system specifically, so it would not be appropriate to compare them with our system. 

We designed the system to detect the wrong-way vehicle from only one side of the road. The camera will focus on only one side. In our future work, we will extend this and implement the system on both sides of the road using a single camera. In that case, the camera will be placed in the middle of the road focusing on both sides.

The system was performed in Python language with OpenCV deep learning library. As the YOLO algorithm is computationally expensive, the frame per second is 2.5 in AMD Ryzen 7 CPU in the windows platform. As it is not necessary to take into account every frame, we took one frame and skip the next four. The GPU version is much faster and by using an Nvidia GPU in the Linux platform, the speed can be increased by several times. 

\section*{CONCLUSIONS}
This paper presented a system that can detect wrong-way vehicles and mark them from on-road video footage. In the first stage, every vehicle in the video frame is detected using the YOLO object detector as it is very accurate and faster than any object detector algorithm. Then, the bounding boxes generated by the YOLO are fed to the centroid based tracking algorithm. The tracking algorithm tracks each vehicle in the specified region of interest. Then, by computing the centroid height of each vehicle in consecutive frames, the direction of vehicles can be determined. The system shows a very promising result as we tested it using traffic footage we collected from the roads. The whole system's accuracy depends on the detection of the YOLO. If YOLO detects correctly, our proposed method will have an accuracy of almost 100\%.

The limitation of this system is with the centroid tracking algorithm. The centroids of the object must lie close together between subsequent frames or the ID number might be switched due to overlap of one object to another. Though this problem doesn’t affect our system much.

\end{document}